\documentclass[conference]{IEEEtran}
\IEEEoverridecommandlockouts
\usepackage{cite}
\usepackage{amsmath,amssymb,amsfonts}
\usepackage{algorithmic}
\usepackage{graphicx}
\usepackage{textcomp}
\usepackage{xcolor}

\newcommand\misscite[1]{\textcolor{orange}{[?]}}

\begin{document}

\title{A system for the 2019 Sentiment, Emotion and Cognitive State Task of DARPA's LORELEI project}
\author{\IEEEauthorblockN{Victor R Martinez\IEEEauthorrefmark{1},
Anil Ramakrishna, Ming-Chang Chiu, Karan Singla and
Shrikanth Narayanan\IEEEauthorrefmark{2}}
\IEEEauthorblockA{Department of Computer Science,\\
University of Southern California\\
\IEEEauthorrefmark{1}victorrm@usc.edu,
\IEEEauthorrefmark{2}shri@sipi.usc.edu}}
\maketitle
%
\begin{abstract}
During the course of a Humanitarian Assistance-Disaster Relief (HADR) crisis, that can happen anywhere in the world, real-time information is often posted online by the people in need of help which, in turn, can be used by different stakeholders involved with management of the crisis. 
Automated processing of such posts can considerably improve the effectiveness of such efforts; for example, understanding the aggregated emotion from affected populations in specific areas may help inform decision-makers on how to best allocate resources for an effective disaster response. 
However, these efforts may be severely limited by the availability of resources for the local language. The ongoing DARPA project \textit{Low Resource Languages for Emergent Incidents} (LORELEI) aims to further language processing technologies for low resource languages in the context of such a humanitarian crisis. 
In this work, we describe our submission for the 2019 Sentiment, Emotion and Cognitive state (SEC) pilot task of the LORELEI project. We describe a collection of sentiment analysis systems included in our submission along with the features extracted.
Our fielded systems obtained the best results in both English and Spanish language evaluations of the SEC pilot task.
\end{abstract}

\section{Introduction}
The growing adoption of online technologies has created new opportunities for emergency information propagation \cite{imran2015processing}. During crises, affected populations post information about what they are experiencing, what they are witnessing, and relate what they hear from other sources \cite{hughes2009twitter}.
This information contributes to the creation and dissemination of situational awareness \cite{vieweg2010microblogging, vieweg2012twitter, imran2014aidr,imran2015processing}, and crisis response agencies such as government departments or public health-care NGOs can make use of these channels to gain insight into the situation as it unfolds \cite{vieweg2010microblogging, vieweg2014integrating}. 
Additionally, these organizations might also post time-sensitive crisis management information to help with resource allocation and provide status reports \cite{harrald2002web}.
While many of these organizations recognize the value of the information found online---specially during the on-set of a crisis---they are in need of automatic tools that locate actionable and tactical information \cite{DBLP:conf/lrec/ImranMC16,imran2015processing}.

Opinion mining and sentiment analysis techniques offer a viable way of addressing these needs, with complementary insights to what keyword searches or topic and event extraction might offer \cite{gaspar2016beyond}.
Studies have shown that sentiment analysis of social media during crises can be useful to support response coordination \cite{purohit2013kind} or provide information about which audiences might be affected by emerging risk events \cite{lachlan2014expressions}.
For example, identifying tweets labeled as ``fear'' might support responders on assessing mental health effects among the affected population \cite{torkildson2014analysis}.
Given the critical and global nature of the HADR events, tools must process information quickly, from a variety of sources and languages, making it easily accessible to first responders and decision makers for damage assessment and to launch relief efforts accordingly \cite{hughes2012evolving, hughes2014online}.
However, research efforts in these tasks are primarily focused on high resource languages such as English, even though such crises may happen anywhere in the world.

The LORELEI program provides a framework for developing and testing systems for real-time humanitarian crises response in the context of low-resource languages. The working scenario is as follows: a sudden state of danger requiring immediate action has been identified in a region which communicates in a low resource language.
Under strict time constraints, participants are expected to build systems that can: translate documents as necessary, identify relevant named entities and identify the underlying situation \cite{Cheung2017ELISASD}.
Situational information is encoded in the form of Situation Frames --- data structures with fields identifying and characterizing the crisis type.
The program's objective is the rapid deployment of systems that can process text or speech audio from a variety of sources, including newscasts, news articles, blogs and social media posts, all in the local language, and populate these Situation Frames. 
While the task of identifying Situation Frames is similar to existing tasks in literature (e.g., slot filling), it is defined by the very limited availability of data \cite{Malandrakis2018}.
This lack of data requires the use of simpler but more robust models and the utilization of transfer learning or data augmentation techniques.

The Sentiment, Emotion, and Cognitive State (SEC) evaluation task was a recent addition to the LORELEI program introduced in 2019, which aims to leverage sentiment information from the incoming documents. This in turn may be used in identifying severity of the crisis in different geographic locations for efficient distribution of the available resources. 
In this work, we describe our systems for targeted sentiment detection for the SEC task. 
Our systems are designed to identify authored expressions of sentiment and emotion towards a HADR crisis. To this end, our models are based on a combination of state-of-the-art sentiment classifiers and simple rule-based systems.
We evaluate our systems as part of the NIST LoREHLT 2019 SEC pilot task.

\section{Previous Work}
\label{sec:prevwork}
Social media has received a lot of attention as a way to understand what people communicate during disasters \cite{schulz2013fine, torkildson2014analysis}.
These communications typically center around collective sense-making \cite{gilles2013collective}, supportive actions \cite{murthy2013twitter,panagiotopoulos2014citizen}, and social sharing of emotions and empathetic concerns for affected individuals \cite{neubaum2014psychosocial}.
To organize and make sense of the sentiment information found in social media, particularly those messages sent during the disaster, several works propose the use of machine learning models (e.g., Support Vector Machines, Naive Bayes, and Neural Networks) trained on a multitude of linguistic features\footnote{For an in-depth review of these approaches, we refer the reader to \cite{beigi2016overview}}.
These features include bag of words, part-of-speech tags, n-grams, and word embeddings; as well as previously validated sentiment lexica such as Linguistic Inquiry and Word Count (LIWC) \cite{pennebaker2015development}, AFINN \cite{IMM2011-06010}, and SentiWordNet \cite{esuli2006sentiwordnet}.
Most of the work is centered around identifying messages expressing sentiment towards a particular situation as a way to distinguish crisis-related posts from irrelevant information \cite{brynielsson2013learning}. Either in a binary fashion (positive vs. negative) (e.g., \cite{brynielsson2013learning}) or over fine-grained emotional classes\footnote{For example, anger, disgust, fear, happiness, sadness, and surprise} (e.g., \cite{schulz2013fine}).

In contrast to social media posts, sentiment analysis of news articles and blogs has received less attention \cite{godbole2007large}. This can be attributed to a more challenging task due to the nature of the domain since, for example, journalists will often refrain from using clearly positive or negative vocabulary when writing news articles \cite{balahur2013sentiment}. However, certain aspects of these communication channels are still apt for sentiment analysis, such as column pieces \cite{kaya2012sentiment} or political news \cite{balahur2013sentiment,de2012media}.

In the context of leveraging the information found online for HADR emergencies, approaches for languages other than English have been limited.
Most of which are done by manually constructing resources for a particular language (e.g., in tweets \cite{ozturk2018sentiment,zielinski2012multilingual, neubig2011safety } and in disaster-related news coverage \cite{sentisail}), or by applying cross-language text categorization to build language-specific models \cite{zielinski2012multilingual, zielinski2013detecting}.

In this work, we develop systems that identify positive and negative sentiments expressed in social media posts, news articles and blogs in the context of a humanitarian emergency. Our systems work for both English and Spanish by using an automatic machine translation system. This makes our approach easily extendable to other languages, bypassing the scalability issues that arise from the need to manually construct lexica resources.



\section{Problem Definition}
This section describes the SEC task in the LORELEI program along with the dataset, evaluation conditions and metrics.

\subsection{The Sentiment, Emotion and Cognitive State (SEC) Task}
\label{sec:sec_task}
Given a dataset of text documents and manually annotated situation frames, the task is to automatically detect sentiment polarity relevant to existing frames and identify the source and target for each sentiment instance. The source is defined as a person or a group of people expressing the sentiment, and can be either a PER/ORG/GPE (person, organization or geo political entity) construct in the frame, the author of the text document, or an entity not explicitly expressed in the document. The target toward which the sentiment is expressed, is either the frame or an entity in the document.

\subsubsection{Situation Frames}
Situation awareness information is encoded into situation frames in the LORELEI program \cite{strassel2016lorelei}. Situation Frames (SF) are similar in nature to those used in Natural Language Understanding (NLU) systems: in essence they are data structures that record information corresponding to a single incident at a single location \cite{Malandrakis2018}. A SF frame includes a situation \textit{Type} taken from a fixed inventory of 11 categories (e.g., medical need, shelter, infrastructure), \textit{Location} where the situation exists (if a location is mentioned) and additional variables highlighting the \textit{Status} of the situation (e.g., entities involved in resolution, time and urgency). An example of a SF can be found in table \ref{tbl:SF}. A list of situation frames and documents serve as input for our sentiment analysis systems.

\begin{table}[]
    \centering
    \caption{Example of a Situation Frame}
    \begin{tabular}{c|p{6.5cm}}
    Original Text & La crisis pol\'itica que comenz\'o en abril pasado en Nicaragua, una situaci\'on in\'edita en la historia reciente del pa\'is, reporta al menos 79 muertos y 868 heridos, seg\'un cifras de la Comisi\'on Interamericana de Derechos Humanos\\
     & (The political crisis that began last April in Nicaragua, a situation unprecedented in the recent history of the country, reports at least 79 deaths and 868 wounded, according to figures from the Inter-American Commission on Human Rights.)\\[3pt]\hline
    \textit{SF-Type} & Medical Need\\
    \textit{Location} & Nicaragua\\\hline
    \textit{Status} & Current, No known resolution, Non-urgent
    \end{tabular}
    \label{tbl:SF}
\end{table}


\subsection{Data}
Training data provided for the task included documents were collected from social media, SMS, news articles, and news wires. This consisted of 76 documents in English and 47 in Spanish. The data are relevant to the HADR domain but are not grounded in a common HADR incident. Each document is annotated for situation frames and associated sentiment by 2 trained annotators from the Linguistic Data Consortium (LDC)\footnote{https://www.ldc.upenn.edu/}.
Sentiment annotations were done at a segment (sentence) level, and included Situation Frame, Polarity (positive / negative), Sentiment Score, Emotion, Source and Target. Sentiment labels were annotated between the values of -3 (very negative) and +3 (very positive) with 0.5 increments excluding 0. Additionally, the presence or absence of three specific emotions: fear, anger, and joy/happiness was marked. If a segment contains sentiment toward more than one target, each will be annotated separately.
Summary of the training data is given in Table \ref{tab:distr}. 

\begin{table}
    \centering
    \caption{Frequency statistics for the provided training data per language: number of documents, number of annotated situation frames, number of sentiment instances, percentage of negative polarity. }
    \begin{tabular}{lcccc}
        & \#Documents & \#SF & \#Sentiment & \% Neg \\
         English & 76 & 85 & 380 & 81.57\\
         Spanish & 47 & 56 & 168 & 98.10\\ \hline
         Total & 123 & 141 & 548 & 84.85\\
    \end{tabular}
    \label{tab:distr}
\end{table}

\subsection{Evaluation}
Systems participating in the task were expected to produce outputs with sentiment polarity, emotion, sentiment source and target, and the supporting segment from the input document. This output is evaluated against a ground truth derived from two or more annotations. For the SEC pilot evaluation, a reference set with dual annotations from two different annotators was provided. The system's performance was measured using variants of precision, recall and f1 score, each modified to take into account the multiple annotations. The modified scoring is as follows: let the agreement between annotators be defined as two annotations with the same sentiment polarity, source, and target. That is, consider two annotators in agreement even if their judgments vary on sentiment values or perceived emotions. Designate those annotations with agreement as ``D'' and those which were not agreed upon as ``S''. When computing precision, recall and f measure, each of the sentiment annotations in D will count as two occurrences in the reference, and likewise a system match on a sentiment annotation in D will count as two matches. Similarly, a match on a sentiment annotation in S will count as a single match. The updated precision, recall and f-measure were defined as follows:

\begin{align*}
    \text{precision} &= \frac{2 * \text{Matches in D} + \text{Matches in S}}{2 * \text{Matches in D} + \text{Matches in S} + \text{Unmatched}}\\[10pt]
    \text{recall} &= \frac{2 * \text{Matches in D} + \text{Matches in S}}{2|D| + |S|}\\[10pt]
    \text{f1} &= \frac{2 * \text{precision} * \text{recall}}{(\text{precision} + \text{recall})}
\end{align*}

\section{Method}
We approach the SEC task, particularly the polarity and emotion identification, as a classification problem. Our systems are based on English, and are extended to other languages via automatic machine translation (to English). In this section we present the linguistic features and describe the models using for the evaluation.

\subsection{Machine Translation}
Automatic translations from Spanish to English were obtained from Microsoft Bing using their publicly available API\footnote{https://www.bing.com/translator}. For the pilot evaluation, we translated all of the Spanish documents into English, and included them as additional training data. 
At this time we do not translate English to Spanish, but plan to explore this thread in future work.

\subsection{Linguistic Features}

\subsubsection{N-grams}
We extract word unigrams and bigrams. These features were then transformed using term frequencies (TF) and Inverse document-frequency (IDF).

\subsubsection{Distributed Semantics}
Word embeddings pretrained on large corpora allow models to efficiently leverage word semantics as well as similarities between words. This can help with vocabulary generalization as models can adapt to words not previously seen in training data. In our feature set we include a 300-dimensional word2vec word representation trained on a large news corpus \cite{mikolov2013distributed}.
We obtain a representation for each segment by averaging the embedding of each word in the segment.
We also experimented with the use of GloVe \cite{pennington2014glove}, and Sent2Vec \cite{pgj2017unsup}, an extension of word2vec for sentences.

\subsubsection{Sentiment Features}
We use two sources of sentiment features: manually constructed lexica, and pre-trained sentiment embeddings.
When available, manually constructed lexica are a useful resource for identifying expressions of sentiment \cite{beigi2016overview}.
We obtained word percentages across 192 lexical categories using Empath\cite{fast2016empath}, which extends popular tools such as the Linguistic Inquiry and Word Count (LIWC) \cite{pennebaker2015development} and General Inquirer (GI) \cite{stone1966general} by adding a wider range of lexical categories. 
These categories include emotion classes such as surprise or disgust.

Neural networks have been shown to capture specific task related subtleties which can complement the manually constructed sentiment lexica described in the previous subsection.
For this work, we learn sentiment representations using a bilateral Long Short-Term Memory model \cite{hochreiter1997long} trained on the Stanford Sentiment Treebank \cite{socher2013recursive}. This model was selected because it provided a good trade off between simplicity and performance on a fine-grained sentiment task, and has been shown to achieve competitive results to the state-of-the-art \cite{DBLP:conf/wassa/BarnesKW17}.

\subsection{Models}
We now describe the models used for this work.
Our models can be broken down into two groups: our first approach explores \textit{state-of-the-art} models in targeted and untargeted sentiment analysis to evaluate their performance in the context of the SEC task.
These models were pre-trained on larger corpora and evaluated directly on the task without any further adaptation.
In a second approach we explore a data augmentation technique based on a proposed simplification of the task.
In this approach, traditional machine learning classifiers were trained to identify which segments contain sentiment towards a SF regardless of sentiment polarity.
For the classifiers, we explored the use of Support Vector Machines and Random Forests.
Model performance was estimated through 10-fold cross validation on the train set. Hyper-parameters, such as of regularization, were selected based on the performance on grid-search using an 10-fold inner-cross validation loop.
After choosing the parameters, models were re-trained on all the available data.

\subsubsection{\textbf{Baselines}}
We consider some of the most popular baseline models in the literature: (i) minority class baseline (due to the heavily imbalanced dataset), (ii) Support Vector Machines trained on TF-IDF bi-gram language model, (iii) and Support Vector Machines trained on word2vec representations. These models were trained using English documents only.

\subsubsection{\textbf{Model I: Pretrained Sentiment Classifiers}}
Two types of targeted sentiment are evaluated for the task: those expressed towards either a situation frame or those towards an entity. To identify sentiment expressed towards an SF, we use the pretrained model described in \cite{DBLP:journals/corr/RadfordJS17}, in which a multiplicative LSTM cell is trained at the character level on a corpus of 82 million Amazon reviews. The model representation is then fed to a logistic regression classifier to predict sentiment. This model (which we will refer to as OpenAI) was chosen since at the time of our system submission it was one of the top three performers on the binary sentiment classification task on the Stanford Sentiment Treebank. 
In our approach, we first map the text associated with the SF annotation with a segment from the document and pass the full segment to the pretrained OpenAI model identify the sentiment polarity for that segment.

To identify sentiment targeted towards an entity, we use the recently released Target-Based Sentiment Analysis (TBSA) model from \cite{li2018unified}. In TBSA, two stacked LSTM cells are trained to predict both sentiment and target boundary tags (e.g., predicting S-POS to indicate the start of the target towards which the author is expressing positive sentiment, I-POS and E-POS to indicate intermediate and end of the target). 
In our submission, since input text documents can be arbitrarily long, we only consider sentences which include a known and relevant entity; these segments are then fed to the TBSA model to predict targeted sentiment. If the target predicted by this model matched with any of the known entities, the system would output the polarity and the target.

\subsubsection{\textbf{Model IIa: Simplifying the Task}}
\label{model:2a}
\label{sec:subtask}
In this model we limit our focus on the task of correctly identifying those segments with sentiment towards a SF. 
That is, given a pair of SF and segment, we train models to identify if this segment contains any sentiment towards that SF. This allows us to expand our dataset from $123$ documents into one with $\sum_d |SF_d| \times |d|$ number of samples, where $|d|$ is the length of the document (i.e., number of segments) and $|SF_d|$ is the number of SF annotations for document $d$. Summary of the training dataset after augmentation is given in Table \ref{tab:distr2}. 

Given the highly skewed label distribution in the training data, a majority of the constructed pairs do not have any sentiment towards a SF. Hence, our resulting dataset has a highly imbalanced distribution which we address by training our models after setting the class weights to be the inverse class frequency. To predict polarity, we assume the majority class of negative sentiment. We base this assumption on the fact that the domain we are working with doesn't seem to support the presence of positive sentiment, as made evident by the highly imbalanced dataset. 

\begin{table}
    \centering
    \caption{Frequency statistics for the train dataset after augmentation}
    \begin{tabular}{lccc}
         & \#(SF$\times$Segments) & With Sentiment & Total\\\hline
         English & 5751 & 285 & 6030\\
         Spanish & 1232 & 132 & 1364\\
    \end{tabular}
    \label{tab:distr2}
\end{table}

\subsubsection{\textbf{Model IIb: Domain-specific models}}
\label{model:2b}
Owing to the nature of the problem domain, there is considerable variance in the source of the text documents and their structure. For example, tweets only have one segment per sample whereas news articles contain an average of $7.07\pm4.96$ and $6.31\pm4.93$ segments for English and Spanish documents respectively. 
Moreover, studies suggest that sentiments expressed in social media tend to differ significantly from those in the news \cite{godbole2007large}.
Table \ref{tab:breakdown} presents a breakdown of the train set for each sentiment across domains, as is evident tweets form a sizeable group of the training set.
Motivated by this, we train different models for tweets and non-tweet documents in order to capture the underlying differences between the data sources.

\begin{table}
    \centering
    \caption{Train dataset domain break-down}
    \begin{tabular}{lcc}
        English & Neg & Pos \\ \hline
         Tweet & 85 & 16 \\
         Others & 204 & 43 \\\hline
         Total & 289 & 59 \\
    \end{tabular}\hspace{1cm}
    \begin{tabular}{lcc}
        Spanish & Neg & Pos \\ \hline
         Tweet & 47 & 1 \\
         Others & 98 & 12 \\\hline
         Total & 145 & 13 \\
    \end{tabular}
    \label{tab:breakdown}
\end{table}

\subsubsection{\textbf{Model IIc: Twitter-only model}}
\label{model:2c}
Initial experiments showed that our main source of error was not being able to correctly identify the supporting segment. Even if polarity, source and target were correctly identified, missing the correct segment was considered an error, and thus lowered our models' precision. To address this, we decided to use a model which only produced results for tweets given that these only contain one segment, making the segment identification sub-task trivial.

\section{Results}
Model performance during train is presented in Table \ref{tab:res}.
While all the models outperformed the baselines, not all of them did so with a significant margin due to the robustness of the baselines selected.
The ones found to be significantly better than the baselines were models IIb (Domain-specific) and IIc (Twitter-only) (permutation test, $n = 10^5$ both $p < 0.05$). The difference in precision between model IIb and IIc points out to the former making the wrong predictions for news articles. These errors are most likely in selecting the wrong supporting segment. Moreover, even though models IIa-c only produce negative labels, they still achieve improved performance over the state-of-the-art systems, highlighting the highly skewed nature of the training dataset.

Table \ref{tab:eval} present the official evaluation results for English and Spanish. Some information is missing since at the time of submission only partial score had been made public. As previously mentioned, the pre-trained \textit{state-of-the-art} models (model I) were directly applied to the evaluation data without any adaptation. These performed reasonably well for the English data. Among the submissions of the SEC Task pilot, our systems outperformed the other competitors for both languages. 

\begin{table}
    \centering
    \caption{Model performance on English train data estimated using 10-fold CV}
    \begin{tabular}{c|c c c}
         Model & Prec & Recall & F1   \\ \hline
         Minority & 0.04 & \textbf{1.00} & 0.08   \\ 
         SVM tfidf & 0.69 & 0.12 & 0.21   \\ 
         SVM W2V & 0.10 & 0.38 & 0.16   \\ \hline
         Model IIa & 0.42 & 0.17 & 0.24   \\ 
         Model IIb & 0.92 & 0.22 & \textbf{0.36}   \\ 
         Model IIc & \textbf{1.00} & 0.22 & \textbf{0.36}   \\ 
    \end{tabular}
    \label{tab:res}
\end{table}

\begin{table}
\centering
\caption{Official Evaluation Results for English and Spanish. Dashes denote missing information (not reported)}
    \begin{tabular}{c|cc}
    \textbf{Polarity}  & Eng & Spa \\ \hline
     Team 1 & 0.33 & 0.02   \\ 
     Team 2 & 0.03 & 0.04    \\ \hline
     Model I & 0.20 & -\\
     Model IIa &0.03 & 0.05\\
     Model IIb &0.32 & 0.35 \\
     Model IIc & \textbf{0.36} &  \textbf{0.39}  \\
    \end{tabular}\\[5pt]
    \label{tab:eval}
\end{table}

\section{Conclusion}
Understanding the expressed sentiment from an affected population during the on-set of a crisis is a particularly difficult task, especially in low-resource scenarios.
There are multiple difficulties beyond the limited amount of data. 
For example, in order to provide decision-makers with actionable and usable information, it is not enough for the system to correctly classify sentiment or emotional state, it also ought to identify the source and target of the expressed sentiment. To provide a sense of trust and accountability on the system's decisions, it makes sense to identify a justifying segment. 
Moreover, these systems should consider a variety of information sources to create a broader and richer picture on how a situation unfolds.
Thus, it is important that systems take into account the possible differences in the way sentiment is expressed in each one of these sources.
In this work, we presented two approaches to the task of providing actionable and useful information.
Our results show that \textit{state-of-the-art} sentiment classifiers can be leveraged out-of-the-box for a reasonable performance on English data.
By identifying possible differences coming from the information sources, as well as by exploiting the information communicated as the situation unfolds, we showed significant performance gains on both English and Spanish.

\bibliographystyle{IEEEtran}
\bibliography{references}

\end{document}